\documentclass[conference]{IEEEtran}
\pdfoutput=1

\usepackage{cite}

\ifCLASSINFOpdf
  \usepackage[pdftex]{graphicx}
  \graphicspath{{./images/}}
  \DeclareGraphicsExtensions{.pdf,.jpeg,.png,.jpg}
\else
\fi
\usepackage{mathtools}
\usepackage{bm} 
\usepackage{calrsfs} 
\usepackage{amsfonts} 
\usepackage{balance}

\usepackage[usenames, dvipsnames]{color}

\ifCLASSOPTIONcompsoc
 \usepackage[caption=false,font=normalsize,labelfont=sf,textfont=sf]{subfig}
\else
 \usepackage[caption=false,font=footnotesize]{subfig}
\fi
\usepackage{url}

\usepackage[colorinlistoftodos]{todonotes} 
\DeclareMathAlphabet{\pazocal}{OMS}{zplm}{m}{n}
\DeclareMathOperator{\diag}{diag}

\usepackage{float}
\newcommand{\qq}[1]{``#1''}

\usepackage{multirow}

\begin{document}
%
\title{Text Classification based on Word Subspace \\with Term-Frequency}



\author{\IEEEauthorblockN{Erica K. Shimomoto$^{*}$, Lincon S. Souza$^{*}$, Bernardo B. Gatto$^{\dagger}$, Kazuhiro Fukui$^{*}$}
\IEEEauthorblockA{
$^{*}$School of Systems and Information Engineering, University of Tsukuba, Japan\\
\{erica,lincons\}@cvlab.cs.tsukuba.ac.jp, kfukui@cs.tsukuba.ac.jp \\
$^{\dagger}$Center for Artificial Intelligence Research (C-Air), University of Tsukuba, Japan\\
bernard.gatto@gmail.com
}
}

\maketitle

\begin{abstract}

Text classification has become indispensable due to the rapid increase of text in digital form. Over the past three decades, efforts have been made to approach this task using various learning algorithms and statistical models based on bag-of-words (BOW) features. Despite its simple implementation, BOW features lack of semantic meaning representation. To solve this problem, neural networks started to be employed to learn word vectors, such as the \textit{word2vec}. \textit{Word2vec} embeds word semantic structure into vectors, where the angle between vectors indicates the meaningful similarity between words. To measure the similarity between texts, we propose the novel concept of \textit{word subspace}, which can represent the intrinsic variability of features in a set of word vectors. Through this concept, it is possible to model text from word vectors while holding semantic information. To incorporate the word frequency directly in the subspace model, we further extend the word subspace to the term-frequency (TF) weighted word subspace. Based on these new concepts, text classification can be performed under the mutual subspace method (MSM) framework. The validity of our modeling is shown through experiments on the Reuters text database, comparing the results to various state-of-art algorithms.

\end{abstract}

%
\IEEEpeerreviewmaketitle

\section{Introduction}
\label{sec:Intro}
Text classification has become an indispensable task due to the rapid growth in the number of texts in digital form available online. It aims to classify different texts, also called documents, into a fixed number of predefined categories, helping to organize data, and making easier for users to find the desired information. Over the past three decades, many methods based on machine learning and statistical models have been applied to perform this task, such as latent semantic analysis (LSA), support vector machines (SVM), and multinomial naive Bayes (MNB).

The first step in utilizing such methods to categorize textual data is to convert the texts into a vector representation. One of the most popular text representation models is the bag-of-words model~\cite{salton1975vector}, which represents each document in a collection as a vector in a vector space. Each dimension of the vectors represents a term (e.g., a word, a sequence of words), and its value encodes a weight, which can be how many times the term occurs in the document. 

Despite showing positive results in tasks such as language modeling and classification~\cite{turney2010frequency,baroni2010distributional,pado2007dependency}, the BOW representation has limitations: first, feature vectors are commonly very high-dimensional, resulting in sparse document representations, which are hard to model due to space and time complexity. Second, BOW does not consider the proximity of words and their position in the text and consequently cannot encode the words semantic meanings.

To solve these problems, neural networks have been employed to learn vector representations of words~\cite{bengio2003neural,collobert2008unified,mnih2009scalable,turian2010word}. In particular, the \textit{word2vec} representation~\cite{mikolov2013efficient} has gained attention. Given a training corpus, \textit{word2vec} can generate a vector for each word in the corpus that encodes its semantic information. These word vectors are distributed in such a way that words from similar contexts are represented by word vectors with high correlation, while words from different contexts are represented by word vectors with low correlation.

One crucial aspect of the \textit{word2vec} representation is that arithmetic and distance calculation between two word vectors can be performed, giving information about their semantic relationship. However, rather than looking at pairs of word vectors, we are interested in studying the relationship between sets of vectors as a whole and, therefore, it is desirable to have a text representation based on a set of these word vectors.

To tackle this problem, we introduce the novel concept of \textit{word subspace}. It is mathematically defined as a low dimensional linear subspace in a word vector space with high dimensionality. Given that words from texts of the same class belong to the same context, it is possible to model word vectors of each class as word subspaces and efficiently compare them in terms of similarity by using canonical angles between the word subspaces. Through this representation, most of the variability of the class is retained. Consequently, a word subspace can effectively and compactly represent the context of the corresponding text. We achieve this framework through the mutual subspace method (MSM)~\cite{fukui2015difference}.

The word subspace of each text class is modeled by applying PCA without data centering to the set of word vectors of the class. When modeling the word subspaces, we assume only one occurrence of each word inside the class. 

However, as seen in the BOW approach, the frequency of words inside a text is an informative feature that should be considered. In order to introduce this feature in the word subspace modeling and enhance its performance, we further extend the concept of word subspace to the term-frequency (TF) weighted word subspace. 

In this extension, we consider a set of weights, which encodes the words frequencies, when performing the PCA. Text classification with TF weighted word subspace can also be performed under the framework of MSM. We show the validity of our modeling through experiments on the  Reuters\footnote{http://www.daviddlewis.com/resources/testcollections/reuters21578} database, an established database for natural language processing tasks. We demonstrate the effectiveness of the word subspace formulation and its extension, comparing our methods' performance to various state-of-art methods.

The main contributions of our work are:
\begin{itemize}
\item The introduction of the concept of word subspace, which
is efficient to represent a text based on the \textit{word2vec} representation.
\item An extension of word subspace to the term-frequency weighted word subspace, which is capable of incorporating word frequency information directly in the subspace model.
\item A comprehensive evaluation of the word subspace concept and its extension, verifying its effectiveness in representing sets of word vectors obtained through the \textit{word2vec}.
\end{itemize}

The remainder of this paper is organized as follows. In Section~\ref{sec:relwork}, we describe the main works related to text classification. In Section~\ref{sec:wordSubspace}, we present the formulation of our proposed word subspace. In Section~\ref{sec:textClassification}, we explain how text classification with word subspaces is performed under the MSM framework. Then, we present the TF weighted word subspace extension in Section~\ref{sec:weigthedSubspace}. Evaluation experiments and their results are described in Section~\ref{sec:exp}. Further discussion is then presented in Section~\ref{sec:discussion}, and our conclusions are described in Section~\ref{sec:conclusion}.

\section{Related Work}
\label{sec:relwork}
In this section, we outline relevant work towards text classification. We start by describing how text data is conventionally represented using the bag-of-words model and then follow to describe the conventional methods utilized in text classification.

\subsection{Text Representation with bag-of-words}
\label{sec:BOW}

The bag-of-words representation comes from the hypothesis that frequencies of words in a document can indicate the relevance of the document to a query~\cite{salton1975vector}, that is, if documents and a query have similar frequencies for the same words, they might have a similar meaning. This representation is based on the vector space model (VSM), that was developed for the SMART information retrieval system~\cite{salton1971smart}. In the VSM, the main idea is that documents in a collection can be represented as a vector in a vector space, where vectors close to each other represent semantically similar documents.

More formally, a document $d$ can be represented by a vector in $\mathbb{R}^{n}$, where each dimension represents a different term. A term can be a single word, constituting the conventional bag-of-words, or combinations of $N$ words, constituting the bag-of-N-grams. If a term occurs in the document, its position in the vector will have a non-zero value, also known as term weight. Two documents in the VSM can be compared to each other by taking the cosine distance between them~\cite{turney2010frequency}.

There are several ways to compute the term weights. Among them, we can highlight some: Binary weights, term-frequency (TF) weights, and term-frequency inverse document-frequency (TF-IDF) weights.

Consider a corpus with documents $D = \{d_i\}_{i=1}^{|D|}$ and a vocabulary with all terms in the corpus $V = \{w_i\}_{i=1}^{|V|}$. The term weights can be defined as:
\begin{itemize}
\item Binary weight: If a term occurs in the document, its weight is 1. Otherwise, it is zero.
\item Term-frequency weight (TF): The weight of a term $w$ is defined by the number of times it occurs in the document $d$.
\begin{equation}
TF(w,d) = n_d^w
\end{equation}

\item Inverse document-frequency: The weight of a term $w$, given the corpus $D$, is defined as the total number of documents $|D|$ divided by the number of documents that have the term $w$, $|D^w|$.
\begin{equation}
IDF(w | D) = \frac{|D|}{|D^w|}
\end{equation}

\item Term-frequency inverse document-frequency (TF-IDF): The weight of a term $w$ is defined by the multiplication of its term-frequency and its inverse document-frequency. When considering only the TF weights, all terms have the same importance among the corpus. By using the IDF weight, words that are more common across all documents in $D$ receive a smaller weight, giving more importance to rare terms in the corpus.
\begin{equation}
TFIDF(w,d | D)=TF \times IDF
\end{equation}
In very large corpus, it is common to consider the logarithm of the IDF in order to dampen its effect.
\begin{equation}
TFIDF(w,d | D)=TF \times log_{10}(IDF)
\end{equation}

\end{itemize}

\subsection{Conventional text classification methods}
\label{sec:textClassification}
\subsubsection{Multi-variate Bernoulli and multinomial naive Bayes}
Multi-variate Bernoulli (MVB) and multinomial naive Bayes (MNB) are two generative models based on the naive Bayes assumption. In other words, they assume that all attributes (e.g., the frequency of each word, the presence or absence of a word) of each text are independent of each other given the context of the class~\cite{mccallum1998comparison}. 

In the MVB model, a document is represented by a vector generated by a bag-of-words with binary weights. In this case, a document can be considered an event, and the presence or the absence of the words to be the attributes of the event. On the other hand, the MNB model represents each document as a vector generated by a bag-of-words with TF weights. Here, the individual word occurrences are considered as events and the document is a collection of word events.

Both these models use the Bayes rule to classify a document. Consider that each document should be classified into one of the classes in $C=\{c_j\}_{j=1}^{|C|}$. The probability of each class given the document is defined as:

\begin{equation}
P(c_j|d_i) = \frac{P(d_i|c_j)P(c_j)}{P(d_i)}.
\end{equation}

The prior $P(d_i)$ is the same for all classes, so to determine the class to which $d_i$ belongs to, the following equation can be used:

\begin{equation}
prediction(d_i) = argmax_{c_j}P(d_i|c_j)P(c_j)
\end{equation}

The prior $P(c_j)$ can be obtained by the following equation:
\begin{equation}
P(c_j) = \frac{1+|D_j|}{|C|+|D|},
\end{equation}

\noindent where $|D_j|$ is the number of documents in class $c_j$.

As for the posterior $P(d_i|c_j)$, different calculations are performed for each model. For MVB, it is defined as:
\begin{equation}
P(d_i|c_j) = \prod_{k=1}^{|V|}P(w_k|c_j)^{t_i^k}(1-P(w_k|c_j))^{1-t_i^k},
\end{equation}

\noindent where $w_k$ is the k-th word in the vocabulary $V$, and $t_i^k$ is the value (0 or 1) of the k-th element of the vector of document $d_i$.

For the MNB, it is defined as:
\begin{equation}
P(d_i|c_j) = P(|d_i|)|d_i|!\prod_{k=1}^{|V|}\frac{P(w_k|c_j)^{n_i^k}}{n_i^k!},
\end{equation}

\noindent where $|d_i|$ is the number of words in document $d_i$ and $n_i^k$ is the k-th element of the vector of document $d_i$ and it represents how many times word $w_k$ occurs in $d_i$.

Finally, the posterior $P(w_k|c_j)$ can be obtained by the following equation:
\begin{equation}
P(w_k|c_j) = \frac{1+|D_j^k|}{|C|+|D|},
\end{equation}

\noindent where $|D_j^k|$ is the number of documents in class $c_j$ that contain the word $w_k$.

In general, MVB tends to perform better than MNB at small vocabulary sizes whereas MNB is more efficient on large vocabularies.

Despite being robust tools for text classification, both these models depend directly on the bag-of-words features and do not naturally work with representations such as \textit{word2vec}.

\smallskip
\subsubsection{Latent Semantic Analysis}
\label{LSA}
Latent semantic analysis (LSA), or latent semantic indexing (LSI), was proposed in~\cite{deerwester1990indexing}, and it extends the vector space model by using singular value decomposition (SVD) to find a set of underlying latent variables which spans the meaning of texts.

It is built from a term-document matrix, in which each row represents a term, and each column represents a document. This matrix can be built by concatenating the vectors of all documents in a corpus, obtained using the bag-of-words model, that is, $ \bm{X} = [ \bm{v}_1, \bm{v}_2, ..., \bm{v}_{|D|}]$, where $\bm{v}_i$ is the vector representation obtained using the bag-of-words model.

In this method, the term-document matrix is decomposed using the singular value decomposition,
\begin{equation}
\bm{X} = \bm{U\Sigma V}^\top,
\end{equation}

\noindent where $U$ and $V$ are orthogonal matrices and correspond to the left singular vectors and right singular vectors of $X$, respectively. $\Sigma$ is a diagonal matrix, and it contains the square roots of the eigenvalues of $X^TX$ and $XX^T$. LSA finds a low-rank approximation of $X$ by selecting only the $k$ largest singular values and its respective  singular vectors,

\begin{equation}
\bm{X}_k = \bm{U}_k\bm{\Sigma}_k \bm{V}_k^{\top}.
\end{equation}

To compare two documents, we project both of them into this lower dimension space and calculate the cosine distance between them. The projection $\bm{\hat{d}}$ of document $\bm{d}$ is obtained by the following equation:
\begin{equation}
\bm{\hat{d}} = \bm{\Sigma}_k^{-1} \bm{U}_k^\top \bm{d}.
\end{equation}

Despite its extensive application on text classification~\cite{ishii2006text,kou2015application,cvitanic2016lda}, this method was initially proposed for document indexing and, therefore, does not encode any class information when modeling the low-rank approximation. To perform classification, 1-nearest neighbor is usually performed, placing a query document into the class of the nearest training document.

\smallskip
\subsubsection{Support Vector Machine}
\label{sec:SVM}
The support vector machine (SVM) was first presented in~\cite{cortes1995support} and performs the separation between samples of two different classes by projecting them onto a higher dimensionality space. It was first applied in text classification by~\cite{joachims1998text} and have since been successfully applied in many tasks related to natural language processing~\cite{wang2012baselines,leopold2002text}.

Consider a training data set $D$, with $n$ samples
\begin{equation}
D = \{(\bm{x}_i,c_i)|\bm{x}_i\in \mathbb{R}^p, c_i \in \{-1,1\} \}_{i=1}^{n},
\end{equation}
where $c_i$ represents the class to which $\bm{x}_i$ belongs to. Each $\bm{x}_i$ is a $p$-dimensional vector. The goal is to find the hyperplane that divides the points from $c_i = 1$ from the points from $c_i = -1$. This hyperplane can be written as a set of points $x$ satisfying:

\begin{equation}
\bm{w} \cdot \bm{x} - b = 0,
\end{equation}
where $\cdot$ denotes the dot product. The vector $\bm{w}$ is perpendicular to the hyperplane. The parameter $\frac{b}{\|\bm{w}\|}$ determines the offset of the hyperplane from the origin along the normal vector $\bm{w}$.

We wish to choose $\bm{w}$ and $b$, so they maximize the distance between the parallel hyperplanes that are as far apart as possible, while still separating the data. 

If the training data is linearly separable, we can select two hyperplanes in a way that there are no points between them and then try to maximize the distance. In other words, minimize $\|\bm{w}\|$ subject to $c_i(\bm{w}\cdot \bm{x}_u-b) \geq 1, i=\{1,2,...,n\}$. If the training data is not linearly separable, the kernel trick can be applied, where every dot product is replaced by a non-linear kernel function.

\section{Word subspace}
\label{sec:wordSubspace}
All methods mentioned above utilize the BOW features to represent a document. Although this representation is simple and powerful, its main problem lies on disregarding the word semantics within a document, where the context and meaning could offer many benefits to the model such as identification of synonyms.

In our formulation, words are represented as vectors in a real-valued feature vector space $\mathbb{R}^{p}$, by using \textit{word2vec}~\cite{mikolov2013efficient}. Through this representation, it is possible to calculate the distance between two words, where words from similar contexts are represented by vectors close to each other, while words from different contexts are represented as far apart vectors. Also, this representation brings the new concept of arithmetic operations between words, where operations such as addition and subtraction carry meaning (eg., \qq{king}-\qq{man}+\qq{woman}=\qq{queen})~\cite{mikolov2013linguistic}.

Consider a set of documents which belong to the same context $D_c = \{d_i\}_{i=1}^{|D_c|}$. Each document $d_i$ is represented by a set of $N_i$ words, $d_i = \{w_k\}_{k=1}^{N_i}$. By considering that all words from documents of the same context belong to the same distribution, a set of words $W_c = \{w_k\}_{k=1}^{N_c}$ with the words in the context $c$ is obtained.

We then translate these words into word vectors using \textit{word2vec}, resulting in a set of word vectors $X_c = \{\bm{x}^k_c\}_{k=1}^{N_c} \in \mathbb{R}^p$. This set of word vectors is modeled into a word subspace, which is a compact, scalable and meaningful representation of the whole set. Such a word subspace is generated by applying PCA to the set of word vectors.

First, we compute an autocorrelation matrix, $\bm{R}_c$:
\begin{equation}
\bm{R}_c = \frac{1}{N_c}\sum_{i=1}^{N_c}\bm{x}^{i}_c\bm{x}_c^{i^{\top}}.
\end{equation}

The orthonormal basis vectors of $m_c$-dimensional subspace $\pazocal{Y}_c$ are obtained as the eigenvectors with the $m_c$ largest eigenvalues of the matrix $\bm{R}_c$. We represent a subspace $\pazocal{Y}_c$ by the matrix $\bm{Y}_c \in \mathbb{R}^{p \times m_c}$, which has the corresponding orthonormal basis vectors as its column vectors.

\section{Text classification based on word subspace}
\label{sec:textClassification}
We formulate our problem as a single label classification problem. Given a set of training documents, which we will refer as corpus, $D = \{d_i\}_{i=1}^{|D|}$, with known classes $C = \{c_j\}_{j=1}^{|C|}$, we wish to classify a query document $d_q$ into one of the classes in $C$. 

Text classification based on word subspace can be performed under the framework of mutual subspace method (MSM). This task involves two different stages: A learning stage, where the word subspace for each class is modeled, and a classification stage, where the word subspace for a query is modeled and compared to the word subspaces of the classes.

In the learning stage, it is assumed that all documents of the same class belong to the same context, resulting in a set of words $W_c = \{w_c^k\}_{k=1}^{N_c}$. This set assumes that each word appears only once in each class. Each set $\{W_c\}_{c=1}^{|C|}$ is then modeled into a word subspace $\pazocal{Y}_c$, as explained in Section~\ref{sec:wordSubspace}.  As the number of words in each class may vary largely, the dimension $m_c$ of each class word subspace is not set to the same value. 

In the classification stage, for a query document $d_q$, it is also assumed that each word occurs only once, generating a subspace $\pazocal{Y}_q$.

To measure the similarity between a class word subspace $\pazocal{Y}_c$ and a query word subspace $\pazocal{Y}_q$, the canonical angles between the two word subspaces are used~\cite{ANGLES}. There are several methods for calculating canonical angles~\cite{hotelling1936relations},~\cite{afriat1957orthogonal}, and~\cite{DBLP:conf/isrr/FukuiY03}, among which the simplest and most practical is the singular value decomposition (SVD). Consider, for example, two subspaces, one from the training data and another from the query, represented as matrices of bases, $\bm{Y}_{c} = [\bm{\Phi}_{1} \ldots \bm{\Phi}_{m_c}] \in \mathbb{R}^{p \times m_c}$ and 
$\bm{Y}_{q} = [\bm{\Psi}_{1} \ldots \bm{\Psi}_{m_q}] \in \mathbb{R}^{p \times m_q}$, where $\bm{\Phi}_{i}$ are the bases for $\pazocal{Y}_c$ and $\bm{\Psi}_{i}$ are the bases for $\pazocal{Y}_q$.
Let the SVD of 
$\bm{Y}_c^{\top}\bm{Y}_q \in \mathbb{R}^{m_c \times m_q}$ be $\bm{Y}_c^{\top}\bm{Y}_q = \bm{U \Sigma V}^{\top}$,
where $\bm{\Sigma} = \diag(\kappa_{1},\ldots,\kappa_{m_c})$, $\{\kappa_{i} \}_{i=1}^{m_q}$ represents the set of singular values. The canonical angles 
$\{\theta_{i} \}_{i=1}^{m_q}$ can be obtained
as $\{\cos^{-1}(\kappa_{1}),\ldots,\allowbreak\cos^{-1}(\kappa_{m_q})\}$ $(\kappa_{1} \geq \ldots \geq \kappa_{m_q})$.
The similarity between the two subspaces is measured by $t$ angles as follows:

\begin{equation}
S_{(\bm{Y}_c,\bm{Y}_q)}[t] = \frac{1}{t}\sum_{i = 1}^{t} \cos^{2} \theta_{i},\; 1 \leq t \leq m_q, \; m_q \leq m_c.
\end{equation}

\begin{figure}[!t]
  \centering
  \includegraphics[width=\columnwidth]{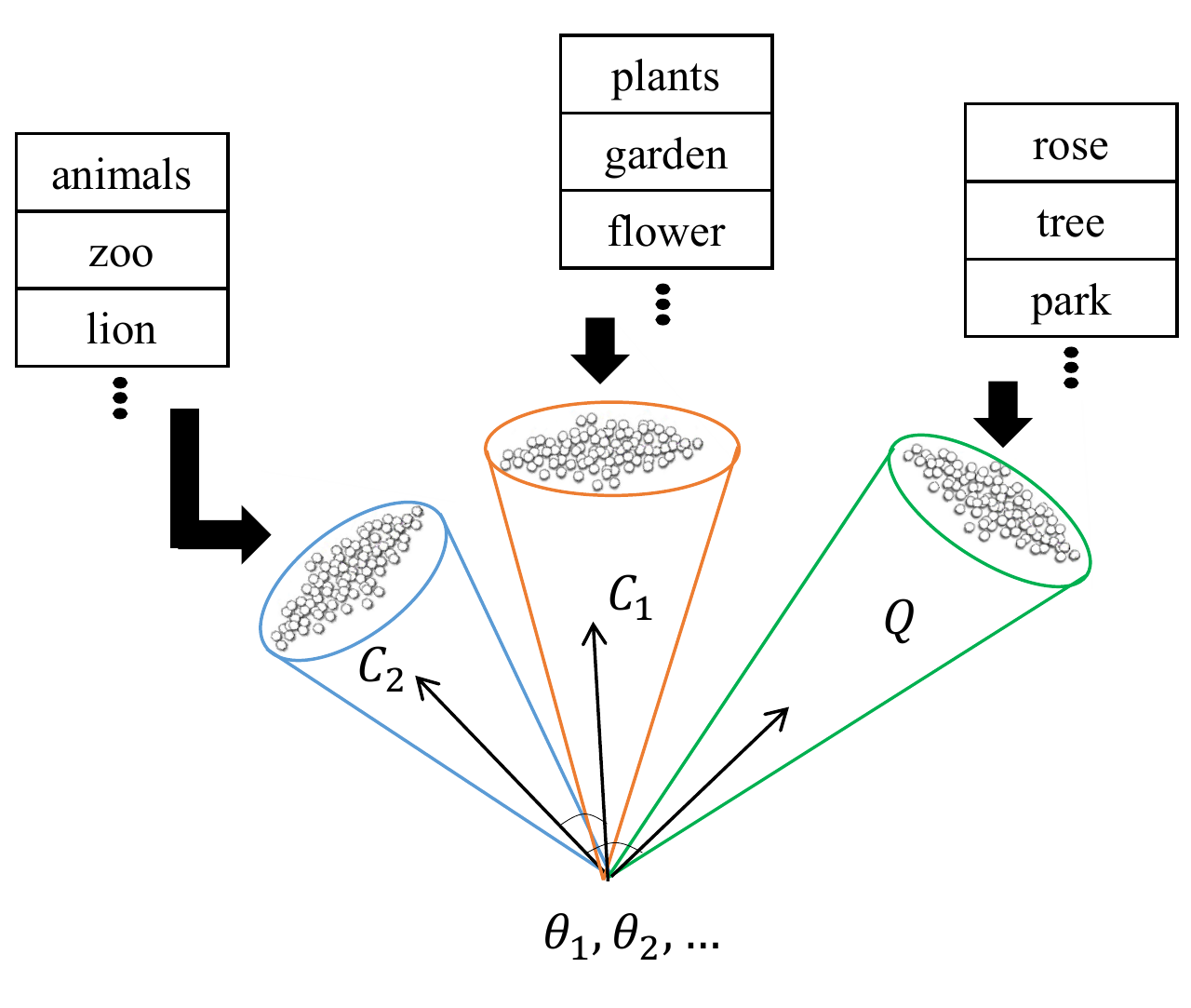}
  \caption{Comparison of sets of word vectors by the mutual subspace method.}
  \label{fig:msm}
\end{figure}

Fig.~\ref{fig:msm} shows the modeling and comparison of sets of words by MSM. This method can compare sets of different sizes, and naturally encodes proximity between sets with related words. 

Finally, the class with the highest similarity with $d_q$ is assigned as the class of $d_q$:
\begin{equation}
prediction(d_q) = argmax_c(S_{(\bm{Y}_c,\bm{Y}_q)}).
\end{equation}

\section{TF weighted word subspace}
\label{sec:weigthedSubspace}
The word subspace formulation presented in Section~\ref{sec:wordSubspace} is a practical and compact way to represent sets of word vectors, retaining most of the variability of features. However, as seen in the BOW features, the frequency of words is relevant information that can improve the characterization of a text. To incorporate this information into the word subspace modeling, we propose an extension of the word subspace, called the term-frequency (TF) weighted word subspace.

Like the word subspace, the TF weighted word subspace is mathematically defined as a low-dimensional linear subspace in a word vector space with high dimensionality. However, a weighted version of the PCA~\cite{greenacre1984theory, jolliffe2006principal} is utilized to incorporate the information given by the frequencies of words (term-frequencies). This TF weighted word subspace is equivalent to the word subspace if we consider all occurrences of the words.

Consider the set of word vectors $\{\bm{x}_c^k\}_{k=1}^{N_c} \in \mathbb{R}^{p}$, which represents each word in the context $c$, and the set of weights $\{\omega_i\}_{i=1}^{N_c}$, which represent the frequencies of the words in the context $c$.

We incorporate these frequencies into the subspace calculation by weighting the data matrix $\bm{X}$ as follows:

\begin{equation}
	\bm{\widetilde{X}}=\bm{X}\bm{\Omega}^{1/2},
\end{equation}

\noindent where $\bm{X} \in \mathbb{R}^{p \times N_c}$ is a matrix containing the word vectors $\{\bm{x}_c^k\}_{k=1}^{N_c}$ and $\bm{\Omega}$ is a diagonal matrix containing the weights $\{\omega_i\}_{i=1}^{N_c}$.

We then perform PCA by solving the SVD of the matrix $\bm{\widetilde{X}}$:
\begin{equation}
	\bm{\widetilde{X}}=\bm{AMB}^{\top},
\end{equation}

\noindent where the columns of the orthogonal matrices $\bm{A}$ and $\bm{B}$ are, respectively, the left-singular vectors and right-singular vectors of the matrix $\bm{\widetilde{X}}$, and the diagonal matrix $\bm{M}$ contains singular values of $\bm{\widetilde{X}}$.


Finally, the orthonormal basis vectors of the $m_c$-dimensional TF weighted subspace $\pazocal{W}$ are the column vectors in $\bm{A}$ corresponding to the $m_c$ largest singular values in $\bm{M}$.

Text classification with TF weighted word subspace can also be performed under the framework of MSM. In this paper, we will refer to MSM with TF weighted word subspace as TF-MSM.

\section{Experimental Evaluation}
\label{sec:exp}

In this section we describe the experiments performed to demonstrate the validity of our proposed method and its extension. We used the Reuters-8 dataset without stop words from \cite{2007:phd-Ana-Cardoso-Cachopo} aiming at single-label classification, which is a preprocessed format of the Reuters-21578\footnote{http://www.daviddlewis.com/resources/testcollections/reuters21578}. Words in the texts were considered as they appeared, without performing stemming or typo correction. This database has eight different classes with the number of samples varying from 51 to over 3000 documents, as can be seen in Table~\ref{tab:Reuters8}.

\begin{table}[!t]
\caption{Document distribution over the classes of the Reuters-8}
\label{tab:Reuters8}
\centering
\begin{tabular}{c | c}
\hline
Class & Number of samples \\
\hline
acq & 2292 \\
crude & 374 \\
earn & 3923 \\
grain & 51 \\
interest & 271 \\
money-fx & 293 \\
ship & 144 \\
trade & 326 \\
\end{tabular}
\end{table}

To obtain the vector representation of words, we used a freely available \textit{word2vec} model\footnote{https://code.google.com/archive/p/word2vec/}, trained by~\cite{mikolov2013efficient}, on approximately 100 billion words, which encodes the vector representation in $\mathbb{R}^{300}$ of over 3 million words from several different languages. Since we decided to focus on English words only, we filtered these vectors to about 800 thousand words, excluding all words with non-roman characters.


To show the validity of our word subspace representation for text classification and the proposed extension, we divided our experiment section into two parts: The first one aims to verify if sets of word vectors are suitable for subspace representation, and the second one puts our methods in practice in a text classification test, comparing our results with the conventional methods described in Section~\ref{sec:relwork}.

\subsection{Evaluation of the word subspace representation}

In this experiment, we modeled the word vectors from each class in the Reuters-8 database into a word subspace. The primary goal is to visualize how much of the text data can be represented by a lower dimensional subspace.

Subspace representations are very efficient in compactly represent data that is close to a normal distribution. This characteristic is due to the application of the PCA, that is optimal to find the direction with the highest variation within the data.

In PCA, the principal components give the directions of maximum variance, while their corresponding eigenvalues give the variance of the data in each of them. Therefore, by observing the distribution of the eigenvalues computed when performing PCA in the modeling of the subspaces, we can suggest if the data is suitable or not for subspace representation.


\begin{figure}[!t]
  \centering
  \includegraphics[width=\columnwidth]{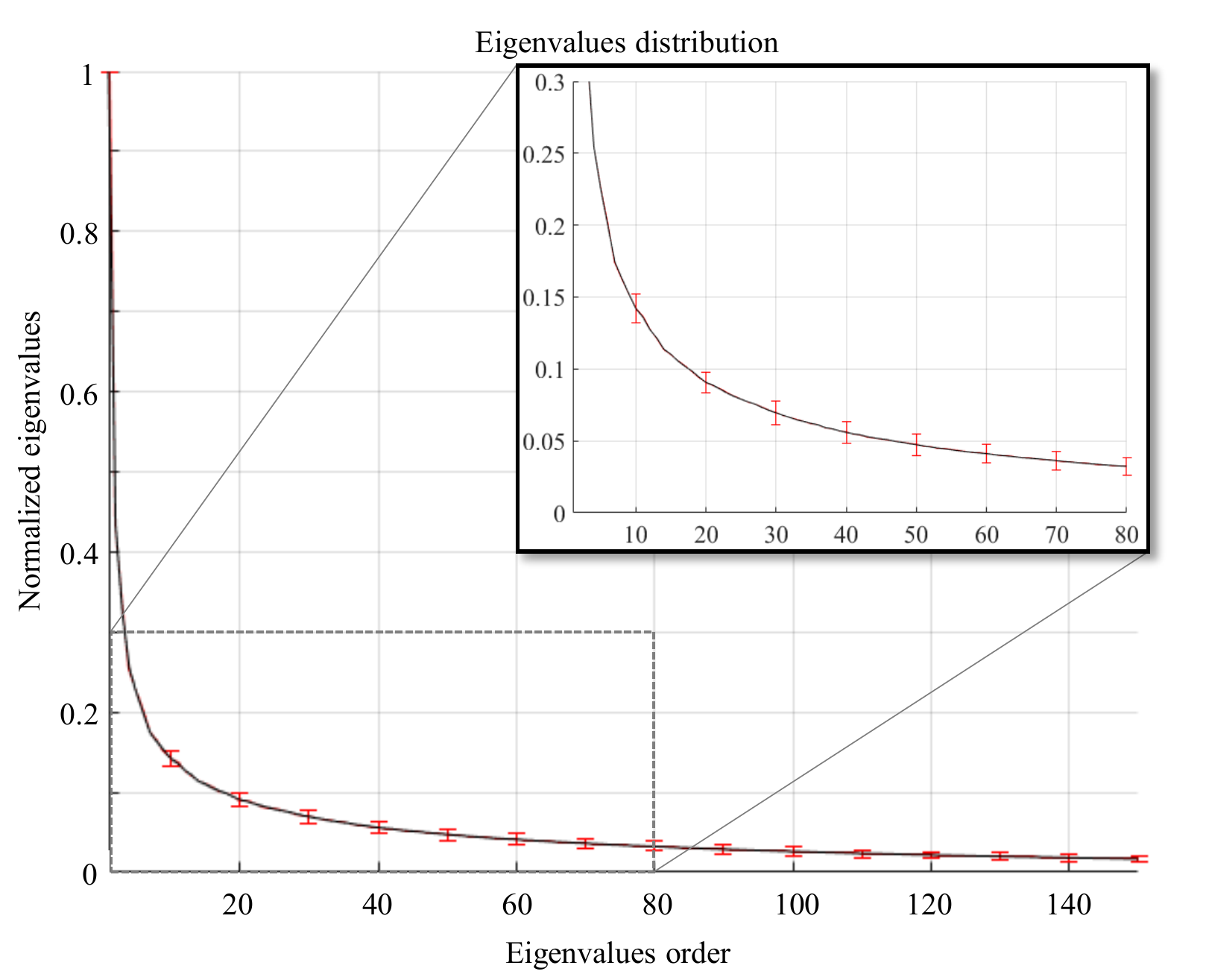}
  \caption{Eigenvalue distribution on word subspaces.}
  \label{fig:eigenvalue}
\end{figure}

For each class, we normalized the eigenvalues by the largest one of the class. Fig.~\ref{fig:eigenvalue} shows the mean of the eigenvalues and the standard deviation among classes. It is possible to see that the first largest eigenvalues retain larger variance than the smallest ones. In fact, looking at the first 150 largest eigenvalues, we can see that they retain, on average, 86.37\% of the data variance. Also, by observing the standard deviation, we can understand that the eigenvalues distribution among classes follows the same pattern, that is, most of the variance is in the first dimensions. This plot indicates that text data represented by vectors generated with \textit{word2vec} is suitable for subspace representation.

\subsection{Text classification experiment}
\label{sec:Exp:TextClassification}
In this experiment, we performed text classification among the classes in the Reuters-8 database. 
We compared the classification using the word subspace, and its weighted extension, based on MSM (to which we will refer as MSM and TF-MSM, respectively) with the baselines presented in Section~\ref{sec:relwork}: MVB, MNB, LSA, and SVM.
Since none of the baseline methods work with vector set classification, we also compared to a simple baseline for comparing sets of vectors, defined as the average of similarities between all vector pair combinations of two given sets. For two matrices $\bm{A}$ and $\bm{B}$, containing the sets of vectors $\{ \bm{x}^{i}_a \}_{i = 1}^{N_A}$ and $\{ \bm{x}^{i}_b \}_{i = 1}^{N_B}$, respectively, where $N_A$ and $N_B$ are the number of main words in each set, the similarity is defined as:
\begin{equation}
Sim_{(A,B)} = \frac{1}{N_A N_B}\sum_{i}^{N_A}\sum_{j}^{N_B}{\bm{x}_a^i}^{\top}\bm{x}_b^j.
\end{equation}
We refer to this baseline as similarity average (SA). For this method, we only considered one occurrence of each word in each set.

Different features were used, depending on the method. Classification with SA, MSM, and TF-MSM was performed using \textit{word2vec} features, to which we refer as w2v. For MVB, due to its nature, only bag-of-words features with binary weights were used (binBOW). For the same reason, we only used bag-of-words features with term-frequency weights (tfBOW) with MNB. Classification with LSA and SVM is usually performed using bag-of-words features and, therefore, we tested with binBOW, tfBOW, and with the term-frequency inverse document-frequency weight, tfidfBOW. We also tested them using \textit{word2vec} vectors. In this case, we considered each word vector from all documents in each class to be a single sample.

To determine the dimensions of the class subspaces and query subspace of MSM and TF-MSM, and the dimension of the approximation performed by LSA, we performed a 10-fold cross validation, wherein each fold, the data were randomly divided into train (60\%), validation (20\%) and test set (20\%).

\begin{table}[!t]
\caption{Results for the text classification experiment on Reuters-8 database}
\label{tab:TextClassification}
\centering
\begin{tabular}{c c | c c}
\hline
 Method & Feature & Accuracy (\%) & Std. Deviation \\
\hline
SA & w2v & 78.73 & 1.56 \\
MSM & w2v & \textbf{90.62} & 0.42 \\
TF-MSM & w2v & \textbf{92.01} & 0.30 \\
\hline
MVB & binBOW &  62.70 & 0.69 \\
MNB & tfBOW & \textbf{91.47} & 0.37 \\
\hline
\multirow{4}{*}{LSA} & w2v & 34.58 & 0.40 \\
					 & binBOW & 86.92 &0.74 \\
					 & tfBOW & 86.23 &	0.96 \\
   					 & tfidfBOW & 86.35 & 1.03 \\
\hline
\multirow{4}{*}{SVM} & w2v & 26.61 & 0.30 \\
				     & binBOW & 89.23 &0.24 \\
					 & tfBOW & 89.10 & 0.29 \\
					 & tfidfBOW & 88.78 & 0.40 \\
\end{tabular}
\end{table}


The results can be seen in Table~\ref{tab:TextClassification}. The simplest baseline, SA with w2v, achieved an accuracy rate of 78.73\%. This result is important because it shows the validity of the \textit{word2vec} representation, performing better than more elaborate methods based on BOW, such as MVB with binBOW.

LSA with BOW features was almost 10\% more accurate than SA, where the best results with binary weights were achieved with an approximation with 130 dimensions, with TF weights were achieved with 50 dimensions, and with TF-IDF weights were achieved with 30 dimensions. SVM with BOW features was about 3\% more accurate than LSA, with binary weights leading to a higher accuracy rate.

It is interesting to note that despite the reasonably high accuracy rates achieved using LSA and SVM with BOW features, they poorly performed when using w2v features.

Among the baselines, the best method was MNB with tfBOW features, with an accuracy of 91.47\%, being the only conventional method to outperform MSM. MSM with w2v had an accuracy rate of 90.62\%, with the best results achieved with word subspace dimensions for the training classes ranging from 150 to 181, and for the query ranging from 3 to 217. Incorporating the frequency information in the subspace modeling resulted in higher accuracy, with TF-MSM achieving 92.01\%, with dimensions of word subspaces for training classes ranging from 150 to 172, and for the query, ranging from 2 to 109. To confirm that TF-MSM is significantly more accurate than MNB, we performed a t-test to compare their results. It resulted in a p-value of 0.031, which shows that at a 95\% significance level, TF-MSM has produced better results.

\section{Discussion}
\label{sec:discussion}

Given the observation of the eigenvalues distribution of word vectors, we could see that word vectors that belong to the same context, i.e., same class, are suitable for subspace representation. Our analysis showed that half of the word vector space dimensions suffice to represent most of the variability of the data in each class of the Reuters-8 database.

The results from the text classification experiment showed that subspace-based methods performed better than the text classification methods discussed in this work. Ultimately, our proposed TF weighted word subspace with MSM surpassed all the other methods.
\textit{word2vec} features are reliable tools to represent the semantic meaning of the words and when treated as sets of word vectors, they are capable of representing the content of texts. 
However, despite the fact that word vectors can be treated separately, conventional methods such as SVM and LSA may not be suitable for text classification using word vectors.

Among the conventional methods, LSA and SVM achieved about 86\% and 89\%, respectively, when using bag-of-words features. Interestingly, both methods had better performance when using binary weights. For LSA, we can see that despite the slight differences in the performance, tfidfBOW required approximations with smaller dimensions. SVM had the lowest accuracy rate when using the tfidfBOW features. One possible explanation for this is that TF-IDF weights are useful when rare words and very frequent words exist in the corpus, giving higher weights for rare words and lower weights for common words. Since we removed the stop words, the most frequent words among the training documents were not considered and, therefore, using TF-IDF weights did not improve the results.

Only MNB with tfBOW performed better than MSM. This result may be because tfBOW features encode the word frequencies, while MSM only considers a single occurrence of words. When incorporating the word frequencies with our TF weighted word subspace, we achieved a higher accuracy of 92.01\%, performing better than MNB at a significance level of 95\%. 


\section{Conclusions and Future Work}
\label{sec:conclusion}

In this paper, we proposed a new method for text classification, based on the novel concept of word subspace under the MSM framework. We also proposed the term-frequency weighted word subspace which can incorporate the frequency of words directly in the modeling of the subspace by using a weighted version of PCA.

Most of the conventional text classification methods are based on the bag-of-words features, which are very simple to compute and had been proved to produce positive results. However, bag-of-words are commonly high dimensional models, with a sparse representation, which is computationally heavy to model. Also, bag-of-words fail to convey the semantic meaning of words inside a text. Due to these problems, neural networks started to be applied to generate a vector representation of words. Despite the fact that these representations can encode the semantic meaning of words, conventional methods do not work well when considering word vectors separately.

In our work, we focused on the \textit{word2vec} representation, which can embed the semantic structure of words, rendering vector angles as a useful metric to show meaningful similarities between words. Our experiments showed that our word subspace modeling along with the MSM outperforms most of the conventional methods. Ultimately, our TF weighted subspace formulation resulted in significantly higher accuracy when compared to all conventional text classification methods discussed in this work.

It is important to note that our method does not consider the order of the words in a text, resulting in a loss of context information. As a future work, we wish to extend our word subspace concept further in mainly two directions. First, we seek to encode word order, which may enrich the representation of context information. Second, we wish to model dynamic context change, enabling analysis of large documents, by having a long-short memory to interpret information using cues from different parts of a text.

\section*{Acknowledgment}
This work is supported by JSPS KAKENHI Grant Number JP16H02842 and the Japanese Ministry of Education, Culture, Sports, Science, and Technology (MEXT) scholarship.




%

\bibliographystyle{IEEEtran}
\balance
\bibliography{bibliography}

\end{document}